\documentclass[final,3p,times,twocolumn]{elsarticle}

\journal{Neurocomputing}

\usepackage{amssymb}
\usepackage{comment}
\usepackage{array}
\usepackage{booktabs}
\usepackage{multirow}
\usepackage{graphicx}
\usepackage{adjustbox}
\usepackage{rotating}
\usepackage{rotating}
\usepackage{siunitx}

\usepackage{amsmath}
\usepackage{hyperref}

\begin{document}

\begin{frontmatter}



\title{Industrial and Medical Anomaly Detection Through \\ Cycle-Consistent Adversarial Networks}

\author[aff1]{Arnaud Bougaham\fnref{fn1}\corref{cor1}}
\ead{arnaud.bougaham@unamur.be}
\author[aff1,aff2]{Valentin Delchevalerie\fnref{fn1}}
\author[aff1]{Mohammed El Adoui}
\author[aff1]{Benoît Frénay}
\fntext[fn1]{ These authors contributed equally to this work.}
\cortext[cor1]{ Corresponding author.}

\affiliation[aff1]{organization={Faculty of Computer Science, NaDI institute},
             addressline={Rue Grandgagnage 21}, 
             city={Namur},
             postcode={5000}, 
             country={Belgium}
}
\affiliation[aff2]{naXys institute}

\begin{abstract}
 In this study, a new Anomaly Detection (AD) approach for industrial and medical images is proposed. This method leverages the theoretical strengths of unsupervised learning and the data availability of both normal and abnormal classes. Indeed, the AD is often formulated as an unsupervised task, implying only normal images during training.
 These normal images are devoted to be reconstructed, through an autoencoder architecture for instance. However, the information contained in abnormal data, when available, is also valuable for this reconstruction. 
 The model would be able to identify its weaknesses by better learning how to transform an abnormal (respectively normal) image into a normal (respectively abnormal) one, helping the entire model to learn better than a single normal to normal reconstruction. 
 To address this challenge, the proposed method uses Cycle-Generative Adversarial Networks (Cycle-GAN) for (ab)normal-to-normal translation.  
 After an input image has been reconstructed by the normal generator, an anomaly score quantifies the differences between the input and its reconstruction. Based on a threshold set to satisfy a business quality constraint, the input image is then flagged as normal or not. The proposed method is evaluated on industrial and medical datasets.
 The results demonstrate accurate performance with a zero false negative constraint compared to state-of-the-art methods. The code is available at \href{https://github.com/ValDelch/CycleGANS-AnomalyDetection}{https://github.com/ValDelch/CycleGANS-AnomalyDetection}.
\end{abstract}


\end{frontmatter}



\section{Introduction}
Image anomaly detection involves identifying anomalies within visual data, playing a crucial role in highlighting unexpected patterns in images.
This work proposes a new approach with a Generative Adversarial Networks (GAN) architecture for the task of Anomaly Detection (AD), which aims to combine the advantages of both unsupervised learning and the data availability of the normal and abnormal classes. Indeed, AD is often formulated as an unsupervised task due to the frequent high imbalance between normal and abnormal data, and the need for generalization across a wide range of anomalies. A common practice is to use an autoencoder architecture to encode / decode normal images. Only the normal class is taken into account, and the reconstruction from the abnormal to the normal class is not included in this training process. Yet, this is precisely the task we are expecting from a reconstruction-based AD method during the inference step. The proposed method seeks to overcome this limitation by learning how to transform an abnormal image into a normal one by exploiting samples from both classes. The objective is to generate a reconstructed image where any abnormal pixel is replaced by a normal one in a visually-coherent manner. During the training step, both a ``normal'' and an ``abnormal'' generator are tied together in an adversarial framework, using Cycle-Generative Adversarial Networks (Cycle-GAN) (proposed by \citet{CycleGAN2017}). Also, reconstructing the abnormal data during the training step yields to a better normal generator than the classical methods using only the normal class. Even if the abnormal datasets can be small as it is usually the case in the AD context, the normal generator performs better, because its performance is also constrained by the abnormal generator, resulting in a good reconstruction. To the best of our knowledge, this is the first time that Cycle-GAN has been studied for this purpose. We still consider this as an unsupervised learning task because the abnormal data used during training is not necessarily representative of all anomalies that could occur. Abnormal data are just given to help during the training phase by giving more feedback to the generators. Therefore, the generalization is guaranteed as it is the case in a classical GAN context, except that the normal reconstruction is less noisy.

\begin{figure}[tb]
    \centering
    \includegraphics[width=1\linewidth]{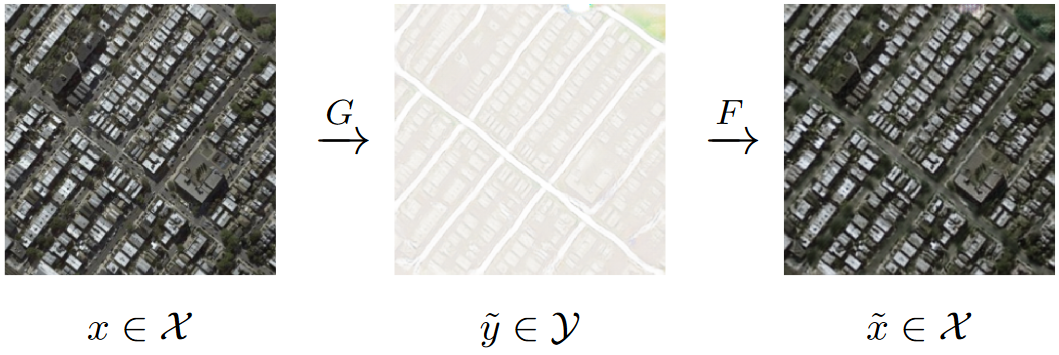}
    \caption{Example generated from a Cycle-GAN (see Section~\ref{sec:experiments} for training details) that learns mappings between aerial photos $\mathcal{X}$ and Google maps $\mathcal{Y}$ (dataset from~\cite{tensorflow2015-whitepaper}). The initial image $x \in \mathcal{X}$ can be mapped to $\tilde{y} \in \mathcal{Y}$ thanks to a first generator $G$. The second one $F$ can then go back from $\tilde{y} \in \mathcal{Y}$ to $\tilde{x} \in \mathcal{X}$. A cycle-consistent constraint enforces $\tilde{x}$ to be close to $x$.}
    \label{fig:intro:toy_example}
\end{figure}

Cycle-GAN is a well-known architecture proposed a few years ago. It constitutes an elegant way to learn conditional mappings from two different domains $\mathcal{X}$ and $\mathcal{Y}$ (for image-to-image translation) by applying a cycle-consistent constraint on the transformations. The popularity of Cycle-GAN lies in the fact that they only need a dataset of unpaired images to learn the mappings. In other words, they do not need the one-to-one correspondence between data from $\mathcal{X}$ and $\mathcal{Y}$ and from $\mathcal{Y}$ and $\mathcal{X}$, but only two independent sets of data $\{x_i \in \mathcal{X}\}$ and $\{y_i \in \mathcal{Y}\}$. As an example, one can consider two unpaired datasets $\{x_i\}$ and $\{y_i\}$ made of unpaired aerial images and Google maps, respectively. A Cycle-GAN can be trained to learn meaningful mappings from $\mathcal{X}$ to $\mathcal{Y}$ and $\mathcal{Y}$ to $\mathcal{X}$. Figure~\ref{fig:intro:toy_example} presents an example generated with this Cycle-GAN.

Numerous studies have demonstrated the versatility of Cycle-GAN in various image analysis applications. Nonetheless, Cycle-GAN remains seldom used in practice to solve problems in the industrial and medical domains. It is for example the case of AD where no prior work directly exploits Cycle-GAN for other purposes than data augmentation. 
Compared to the state-of-the-art methods for AD in images as shown by~\cite{akcay_ganomaly_2019},~\cite{defard_padim_2021} or~\cite{Roth_2022_CVPR}, Cycle-GAN seems to be more suitable than concurrent methods that rely on traditional methods.
Furthermore, we show that the formalism behind Cycle-GANs makes them efficient and well-suited for AD. Specifically, it introduces an identity loss for the reconstruction of normal-to-normal (and abnormal-to-abnormal) images. This makes the generators much less noisy than traditional GANs, making it possible to better discriminate normal and abnormal images. To illustrate this, we focus our experiments on several industrial and medical problems of AD. This is motivated by~(i) the abundance of AD problems in these domains, and~(ii) the positive societal impact of developing efficient AD algorithms for them. \\

The main contributions of our work could be listed as follow:
\begin{itemize}
    \item Use abnormal data in the training process by leveraging a Cycle-GAN architecture for AD, thus considering an identity loss that allows a better discrimination between normal and abnormal images.
    \item Characterize and discuss the performances of the method for diverse industrial and medical AD problems.
    \item Conduct an extensive benchmark to compare the proposed approach with state-of-the-art methods.
    \item Discuss why Cycle-GAN is well-suited for AD in specific image types and explore its potential in industrial and medical domains.
\end{itemize}

In the following, Section~\ref{sec:background_CycleGANS} presents the theoretical prerequisites to understand Cycle-GAN. After that, Section~\ref{sec:related_works} presents the previous works, and highlights that most of them only use simple architectures up to the traditional GAN ones, by training with only normal data. The proposed AD method with Cycle-GAN is described in Section~\ref{sec:methods}. Section~\ref{sec:experiments} then introduces the considered datasets, the experimental setup as well as the results. Finally a discussion is presented in Section~\ref{sec:discussion_limitations} before concluding with Section~\ref{sec:conslusion}.

\section{Background on Cycle-GAN}\label{sec:background_CycleGANS}

This section introduces the formalism for Cycle-Generative Adversarial Network (Cycle-GAN); the building blocks and the loss functions are described.

\subsection{Building Blocks}

Cycle-GAN learns image-to-image mappings from an unpaired dataset composed of two types of images from domains $\mathcal{X}$ and $\mathcal{Y}$. Cycle-GAN is obtained by tying together two distinct conditional GANs with a cycle-consistent constraint. The first GAN is made of a generator 
\begin{equation}
    G: \mathcal{X} \cup \mathcal{Y} \longrightarrow \mathcal{Y}: G\left(z\right)=\tilde{y}, \nonumber
\end{equation}
and a discriminator $D_Y(\cdot)$, and the other is made of a generator
\begin{equation}
    F: \mathcal{X} \cup \mathcal{Y} \longrightarrow \mathcal{X}: F\left(z\right)=\tilde{x}, \nonumber
\end{equation}
and a discriminator $D_X(\cdot)$. For convenience, let us already consider an AD task where $\mathcal{X}$ are abnormal images, while $\mathcal{Y}$ are normal ones. On the one hand, the aim of $G$ is to generate from $x \in \mathcal{X} \cup \mathcal{Y}$ an image such that $D_Y$ cannot distinguish it from real normal images in $\mathcal{Y}$. On the other hand, $F$ aims to generate images such that $D_X$ is fooled and cannot distinguish it from real abnormal images in $\mathcal{X}$. To achieve this, Cycle-GAN is trained with a combination of different losses that are described in the next section.

\subsection{Objective Function}

A Cycle-GAN is made of two GANs that are tied together with a cycle-consistent constraint. The loss can be broken down into three parts so that
\begin{equation}
    G^*, F^* = \text{arg} \min_{F, G} \max_{D_X, D_Y} \mathcal{L}_{\text{adv}} + \lambda_{\text{cyc}} \mathcal{L}_{\text{cyc}} + \lambda_{\text{ide}} \mathcal{L}_{\text{ide}},
    \label{eq:loss_cgan}
\end{equation}
where $\lambda_{\text{cyc}}$ and $\lambda_{\text{ide}}$ are meta-parameters that constraint the different parts of the loss.

The first part of~\eqref{eq:loss_cgan} is made of two classical adversarial losses~\cite{gan}
\begin{equation}
    \mathcal{L}_{\text{adv}} = \mathcal{L}_{\text{GAN}}\left(G, D_Y\right) + \mathcal{L}_{\text{GAN}}\left(F, D_X\right), \nonumber
\end{equation}
where,
\begin{align}
    \mathcal{L}_{\text{GAN}}\left(G, D\right) &= \mathbb{E}_y\left[\log \left(D\left(y\right)\right)\right] \nonumber \\
    &+ \mathbb{E}_x\left[\log \left(1-D\left(G\left(x\right)\right)\right)\right]. \nonumber
\end{align}
On the one side, by enforcing $G$ (resp. $F$) to minimize $\mathcal{L}_\text{adv}$, the generator will try to generate images that look similar to images from $\mathcal{Y}$ (resp. $\mathcal{X}$). On the other side, by enforcing $D_Y$ (resp. $D_X$) to maximize $\mathcal{L}_\text{adv}$, the discriminator will try to distinguish between images coming from the generator $G$ (resp. $F$) and real images in $\mathcal{Y}$ (resp. $\mathcal{X}$).

The second part of~\eqref{eq:loss_cgan} is motivated by the fact that the reconstructed images $F\left(G\left(x\right)\right)$ and $G\left(F\left(y\right)\right)$ should be close to $x$ and $y$, respectively. In other words, the pair of GANs should be cycle-consistent. This is achieved by the cycle-consistent loss
\begin{equation}
    \mathcal{L}_{\text{cyc}} = \mathbb{E}_x\left[ \| F\left(G\left(x\right)\right) - x \|_1 \right] + \mathbb{E}_y\left[ \| G\left(F\left(y\right)\right) - y \|_1 \right], \nonumber
\end{equation}
where a L1 norm is used in the original work on Cycle-GAN~\cite{CycleGAN2017}.

In addition to the $\mathcal{L}_{\text{adv}}$ and $\mathcal{L}_{\text{cyc}}$, an identity loss is added to constrain the generators to leave the images unmodified if they are already in the desired output domain, defined as
\begin{equation}
    \mathcal{L}_{\text{ide}} = \mathbb{E}_x\left[ \| F\left(x\right) - x \|_1 \right] + \mathbb{E}_y\left[ \| G\left(y\right) - y \|_1 \right], \nonumber
\end{equation}
so as to enforce $F\left(x\right)=x$ and $G\left(y\right)=y$. In other words, $F$ should not add anomalies if the input image is already abnormal, and $G$ should not make any modification if it is already normal. Although the identity loss is present in the implementation of Cycle-GAN from the original paper, it is not discussed and seldom used in practice. However, in the context of AD, the use of the identity loss is particularly relevant. Indeed, it is expected from $G$ that it erases any abnormal pixel from the image. Nonetheless, in the case the image does not contain any of them, it should learn to leave it unmodified. This important property is enforced by the identity loss. 

\section{Related Works}\label{sec:related_works}

AD has long been an area of great concern in a wide range of fields such as biomedicine~\cite{schlegl2017unsupervised}, industry~\cite{bougaham2021ganodip} and security~\cite{kiran2018overview, abdallah2016fraud}. Furthermore, a significant number of works have been published to characterize the AD approaches in the literature. The scope of this section is focused on previous works based mainly on GANs and Cycle-GAN, applied to the industrial and medical domains. GANs are used for many image-related tasks, specifically in AD~\cite{schlegl_f-anogan_2019, akcay_ganomaly_2019, zenati_efficient_2019, bougaham2023composite}, segmentation, data augmentation, etc. 
However, in the AD context, Cycle-GAN has been mostly used for data augmentation only~\cite{pandey2020image, kerepecky2022dual, dirvanauskas2019hemigen}. Since our research is focused on AD in medical and industrial applications, in this section, we review the most relevant researches applied to these two fields.
An in-depth analysis of the state-of-the-art methods associated with our research issue shows that recent AD methods are mainly using only GANs. Yet, Cycle-GAN can be highly useful for AD thanks to the combination of unsupervised learning and the data availability of the normal and abnormal classes.

Regarding the industrial studies, \citet{bougaham2021ganodip} propose to use intermediate patches (i.e., parts of image) for the inference step after a Wasserstein GAN training process. The objective is to produce an efficient approach for AD on real industrial images of electronic Printed Circuit Board Assembly (PCBA). The technique can be used to assist current industrial image processing algorithms and to avoid tedious manual processing. Nevertheless, due to the wide variety of possible anomalies in a PCBA and the high complexity of autoencoder architectures, a real-world implementation remains a challenging task, specifically for small anomalies, even if the method evolved to overcome some limitations in the work of \citet{bougaham2023composite}. \citet{akcay_ganomaly_2019} use an autoencoder with skip connections, culminating in a GAN discriminator, which proves to be an effective means of training the model for the normal class, in the context of AD. Different from GAN techniques, \citet{Roth_2022_CVPR} recently suggest a patch-features encoding AD method applied to industrial images, modifying the prediction threshold to ensure a 100\% recall rate with no false negatives. Also, \citet{defard_padim_2021} use a pre-trained CNN for patch embedding and employ multivariate Gaussian distributions to obtain a probabilistic representation of the normal class. \citet{defect_synthesis_1} and \citet{defect_synthesis_2}, as for them, suggest to use Cycle-GAN to perform data augmentation by generating synthetic images for industrial inspection. 

Regarding the medical field, \citet{schlegl_f-anogan_2019} proposed an unsupervised AD framework GANs (f-AnoGAN), that can detect the unseen anomalies of medical subjects after being trained on healthy tomography images. Among the previous studies in medical imaging, works of \citet{hammami2020cycle} and \citet{sandfort2019data} could be cited. These authors use Cycle-GAN to perform the data augmentation task for Magnetic Resonance Imaging (MRI) and Computed Tomography (CT) scan images, respectively. Again, they show that using Cycle-GAN for data augmentation leads to better segmentation performances afterwards.

Despite being better than the previous approaches, the above AD approaches use unsupervised deep learning techniques, such as autoencoders, pre-trained CNN and GANs, to characterize the normal class, without using the insights given by the anomalies. In contrast, in this work, the anomaly images are leveraged immediately in the training phase, using it as prior knowledge to strengthen the model at recognizing anomalies. Furthermore, we assess our method in both industrial and medical images while enforcing zero false negative (ZFN), which is the most useful in these fields where missed detections have large impacts on customers or patients.

In a concise way, a thorough analysis of the most important studies in the literature shows that in the industrial and medical domains, Cycle-GAN has been mainly used for data augmentation. This paper demonstrates the suitability of Cycle-GAN for AD, in particular for industrial and medical images, which has never been covered in the literature. 

\section{Methods}\label{sec:methods}

\begin{figure*}[h]
    \centering
    \includegraphics[width=1\linewidth]{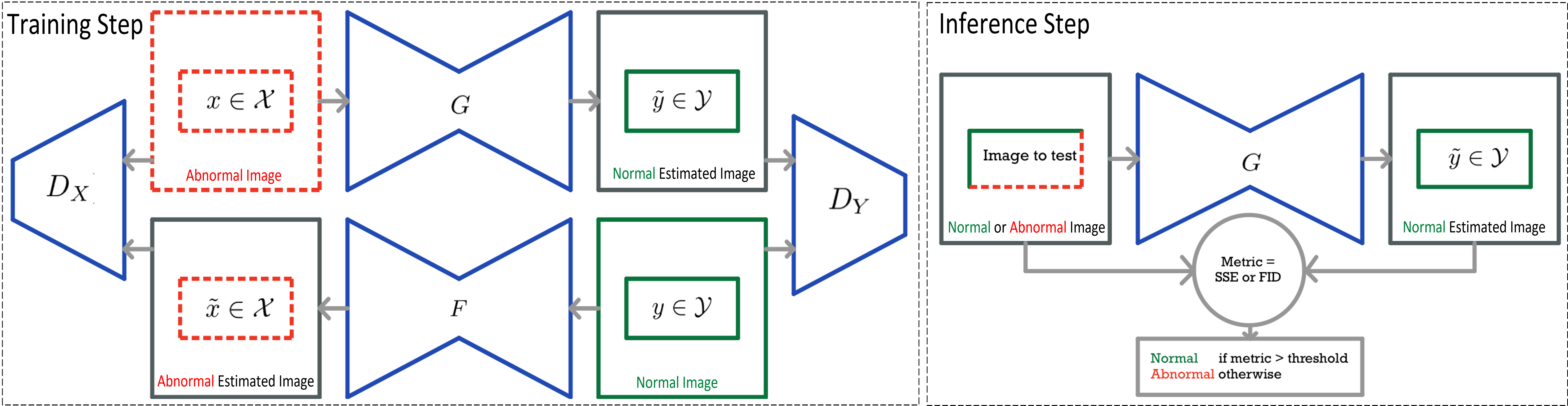}
    \caption{(Color online) Architecture for the training (left side) and the inference (right side) steps (inspired from~\cite{CycleGAN2017}). During the training step, the first generator $G$ tries to map abnormal to normal images by fooling the discriminator $D_Y$ that should not detect fake images. $F$ and $D_X$ follow the same idea but for normal images as input. During the inference step, only $G$ is used even if the input can either be normal or abnormal. 
    }
    \label{fig:Architecture}
\end{figure*}

This section introduces the developed approach and shows its relevance for AD with Cycle-GAN. The training and inference steps are illustrated in Figure~\ref{fig:Architecture}.

The basic idea behind the use of Cycle-GAN for AD is to exploit the conditional mapping learned by one of the two generators: the one that goes from abnormal to normal images. Indeed, by forward-propagating an abnormal image in this generator, it is expected to obtain a new image where the anomaly is erased. Nonetheless, and thanks to the identity loss, if a normal image is forward-propagated in the generator, it is expected to remain unchanged. Therefore, by comparing the output of the generator with its input, anomalies in the input images can be located. The other generator (normal-to-abnormal) is not used for AD. It is only useful to jointly train the first one, similarly to the two discriminators, but not for the AD inference step.

To perform AD, the normal and abnormal test images are given to the learned abnormal-to-normal generator.
Then, an anomaly score is computed to measure the distance between the original test image and the reconstructed one. In this paper, two metrics are considered: a per-pixel sum of the squared differences (SSE), and a Frechet Inception Distance (FID)~\cite{heusel2017gans}. The FID anomaly score is more elaborated and focuses on perceptual differences thanks to the use of a pre-trained Inception V3 network~\cite{inceptionV3}. 
Two different thresholds are considered for the anomaly detector.
The first threshold is set by minimizing the number of classification errors, which yields an anomaly detector with maximum accuracy (ACC). The second one is set so that all true positives are detected (only false alarms can be raised, but no anomaly can be missed). This setting yields an anomaly detector with zero false negative (ZFN), which is the most useful in the critical business applications of the industrial or medical fields, where false negatives have large consequences for customers or patients. In summary, four anomaly detectors are built, i.e., one for each pair of metrics and thresholds. In addition, the receiver operating characteristic curve (AUCROC)~\cite{bradley1997use} is also considered, in order to quantify the performance at different threhsolds. Notice that the thresholds are set on the test sets, and make it possible to assess how much the two distributions (SSE and FID for normal and abnormal data) are discriminated. The use of an additional validation set should be preferred, but it would have been too costly in terms of abnormal data for several datasets (due to the scarcity of abnormal data). Therefore, the accuracy values are an overestimate of the classification performances, and should be seen as a metric to quantify how much the method highlights abnormal images compared to normal images on the test sets. Our goal is indeed to measure the discriminative power of Cycle-GAN for AD, so as to prospectively validate the practical interest of our idea in the industrial and medical domains. 

\section{Experiments}\label{sec:experiments}

This section presents the experiments carried out to evaluate the proposed AD method. First, the datasets and the data preprocessing steps are introduced. Next, the model architecture is detailed. Finally, qualitative and quantitative results are presented, and then discussed in Section~\ref{sec:discussion_limitations}.

\subsection{Datasets}

Seven datasets are used for both the industrial and medical domains, where several types of anomalies may occur. To assess the strengths and weaknesses of our method, we defined four categories of anomalies: small / large object-shaped, or small / large textured-shaped anomalies. Indeed, the anomaly can either arise on a specific object characterized by abrupt changes in pixels intensity (a picture of a screw with a broken head for instance), or where the pixels intensity changes are more progressive and more homogeneous (an abnormal color in a wood picture for instance).

To cover the industrial side, the public MVTEC-AD dataset~\cite{bergmann2019mvtec} is used. It consists of different high resolution industrial images from 15 different categories of object and texture-shaped products with and without anomalies. In this work, 4 datasets were selected from MVTEC-AD to cover the different natures of images: the Hazelnut (large object-shaped images), the Screw (small object-shaped images), the Tile (large texture-shaped images) and the Wood (small texture-shaped images) dataset, which are made of $501$, $480$, $347$ and $326$ images, respectively. All of these datasets are clearly imbalanced with a minority of abnormal images.

To investigate the medical side, three datasets of object and texture-shaped images are used, coming from healthy and unhealthy subjects. 
The first dataset is made of $253$ Brain MRI images (large object-shaped images)~\cite{brainMRI}. Second, Breast Ultrasound (large texture-shaped images)~\cite{breast_tumor} is made of $789$ images. One should mention that for this dataset, many images were manually labeled by experts by highlighting the tumor on the images. Therefore, to avoid any bias during training, we removed all those annotated images from the dataset, resulting in a dataset of $654$ images. Finally, the Retina OCT dataset (small and large texture-shaped anomalies)~\cite{KERMANY20181122} contains $83,600$ images of Optical Coherence Tomography. All of these medical datasets are imbalanced with a minority of normal images.
Table~\ref{table:datasets_table} provides an overview of the number of normal and abnormal images for each dataset.

\begin{table*}[h]
    \renewcommand{\arraystretch}{1}
    \begin{minipage}{\textwidth} 
    \caption{Sizes of the training and test sets for each dataset, regarding the number of normal and abnormal data. A particular split to guarantee that the test sets are balanced is always chosen, even if the initial datasets are imbalanced.}\label{table:datasets_table}
    {\setlength{\tabcolsep}{0pt}
    \begin{center}
        \begin{tabular*}{\textwidth}{@{\extracolsep{\fill}}lcccc@{\extracolsep{\fill}}}
            \toprule%
            & \multicolumn{2}{@{}c@{}}{\textit{Training set}} & \multicolumn{2}{@{}c@{}}{\textit{Test set}}
            \\\cmidrule{2-3}\cmidrule{4-5}%
            \textit{Dataset} & \textit{\# Normal} & \textit{\# Abnormal} & \textit{\# Normal} & \textit{\# Abnormal}
            \\\cmidrule{2-2}\cmidrule{3-3}\cmidrule{4-4}\cmidrule{5-5}%
            Hazelnut & 396 & 35 & 35 & 35 \\
            Screw & 301 & 59 & 60 & 60 \\
            Tile & 222 & 43 & 41 & 41 \\
            Wood & 236 & 30 & 30 & 30 \\
            Brain MRI & 49 & 106 & 49 & 49 \\
            Breast Ultrasound & 67 & 455 & 66 & 66 \\
            Retina OCT & 13,172 & 44,084 & 13,172 & 13,172 \\
            \bottomrule%
        \end{tabular*}
    \end{center}}
    \end{minipage}
\end{table*}

Some of the aforementioned datasets come with different types of anomalies. In this case, a single abnormal class is created by aggregating all the abnormal classes together. Also, the normal and abnormal classes are sometimes imbalanced. Therefore, in order to avoid the use of an imbalanced metric to evaluate the results, a specific split of the dataset is applied to ensure that the test sets are fully balanced. This is obtained by keeping half of the minority class images for the test set, as well as the same number of randomly picked images from the majority class. All the remaining images are left to the training set.
Given that AD may sometimes be a very imbalanced predictive problem, one of the two classes is generally overpopulated in the training sets. However, the test sets are perfectly balanced, which allows us to assess the performances with a simple accuracy metric. Furthermore, even if using imbalanced training sets may hurt the performances for most of the supervised machine learning algorithms, the training process of Cycle-GAN is less sensitive to this. The task of Cycle-GAN differs from a simple label prediction, and each image gives feedback to the two generators, either directly or indirectly. 
The same data preprocessing steps are performed for all the datasets. Images are resized to a resolution of $256 \times 256$ pixels by using a bicubic interpolation method.
Data augmentation is also performed so that objects and textures are rotated and flipped along both axes (except for Retina OCT where only flipping along the horizontal axis is pertinent).

\subsection{Network Architecture and Training Procedure}

For convenience and practical purposes, the architectures used in this work as well as the training procedures are similar for the different applications, and follow the experimental setup presented in the initial paper on Cycle-GAN~\cite{CycleGAN2017}. The generators are formed by three convolution layers, several residual blocks~\cite{residual_blocks}, two fractionally-strided convolution layers and one final convolution layer. We use 9 residual blocks for images resized at $256 \times 256$ resolution, and instance normalization.
For the discriminators, we use $70 \times 70$ PatchGANs~\cite{patchgan1,patchgan2,patchgan3}. All the models are trained through 200 iterations of the Adam optimizer with a learning rate of $2 \times 10^{-4}$.  
A linear learning decay is introduced at the middle of the training. The meta-parameters $\lambda_{\text{cyc}}$ and $\lambda_{\text{ide}}$ are fixed to 10 and 5, respectively. To give an idea of the computation time, training the Cycle-GAN on 500 images (4500 images with the data augmentation) with a $256 \times 256$ resolution for 200 iterations roughly takes 12 hours on a single Nvidia A100 GPU. All experiments have been performed with CUDA 11.3 and Pytorch 1.10.2.

\subsection{Experimental Results}

This subsection showcases the qualitative and quantitative assessments.

\subsubsection{Qualitative Assessments}

\begin{figure*}
    \centering
     \includegraphics[width=0.95\linewidth]{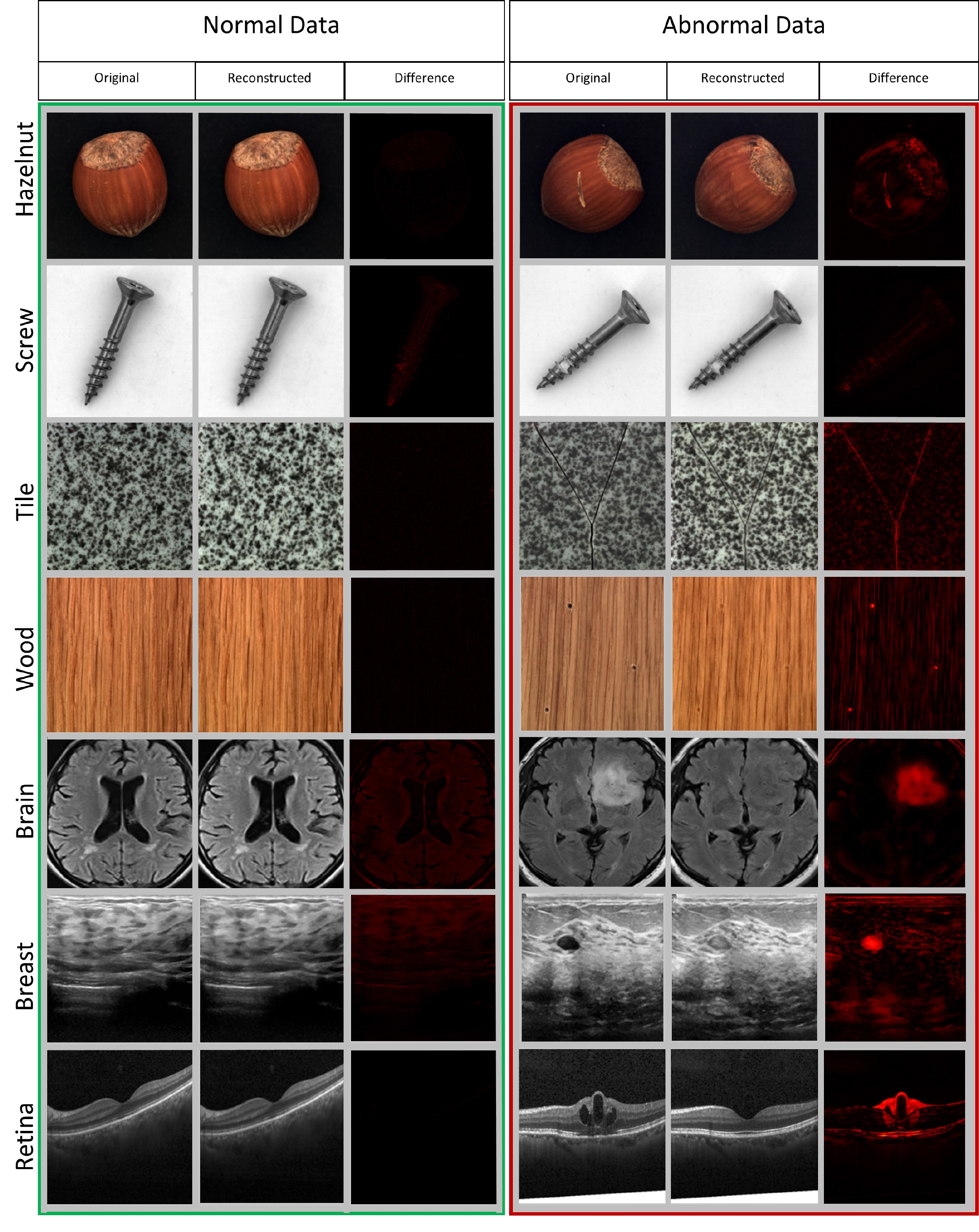}
    \caption{(Color online) Industrial and medical image examples. For each dataset, the left green-framed block presents
    normal images and the right red-framed block shows abnormal images, with the original image (1st column), the normal version generated (2nd column), and their squared pixel-wise difference image (3rd column).}
     \label{fig:Quali}
\end{figure*}

The quality of the reconstruction, as well as the highlighting of anomalies are presented in Figure~\ref{fig:Quali}. It shows the original image, the normal (generated) version and their squared pixel difference image, for the selected industrial and medical datasets.
Notice that the difference images (3rd column) of the abnormal data highlight the anomaly, unlike the ones of the normal data. These images have been specifically chosen to illustrate different typical cases. The following quantitative assessment evaluates the global performances on all the test sets, which are in agreement with the qualitative examples presented below.

\subsubsection{Quantitative Assessments}

Table~\ref{Metric_table} shows the quantitative results of our method (CycleGAN-AD-256). For comparison purposes, the same architecture is assessed with a lower input image resolution of $64 \times 64$, and only 6 residual blocks for the generator (CycleGAN-AD-64). Three different methods are also compared to ours, namely Ganomaly~\cite{akcay_ganomaly_2019}, Padim~\cite{defard_padim_2021} and PatchCore~\cite{Roth_2022_CVPR}.

\begin{sidewaystable*}
    \centering
    \caption{Quantitative Performance Metrics for all the different models on all the different industrial (4 first) and medical (3 last) datasets. For each model, the accuracy with the zero false negative constraint (ZFN), the maximum accuracy (ACC) and the AUCROC (AUC) metrics are presented, considering FID (when applicable) and SSE metrics. All the metrics are the mean ± the standard deviation (both in percent) over 5 runs for which random train and test splits were generated. The mean performance is computed for each model in the last line w.r.t. their most profitable setting (SSE or FID) when available, and bold values are the best of each metric.}
    \vspace{1cm}
    \begin{adjustbox}{width=\textheight}
    \renewcommand{\arraystretch}{3}
        \begin{tabular}{|c|c|c|c|c||c|c|c||c|c|c||c|c|c||c|c|c|c|}
        \cline{3-17}
        \multicolumn{1}{c}{} & \multicolumn{1}{c|}{} & \multicolumn{3}{c||}{\textbf{CycleGAN-AD-256} \textit{(ours)}} & \multicolumn{3}{c||}{\textbf{CycleGAN-AD-64} \textit{(ours)}} & \multicolumn{3}{c||}{\textbf{Ganomaly}~\cite{akcay_ganomaly_2019}} & \multicolumn{3}{c||}{\textbf{Padim}~\cite{defard_padim_2021}} & \multicolumn{3}{c|}{\textbf{PatchCore}~\cite{roth2022towards}}\\
        \cline{3-17}
        \multicolumn{1}{c}{} & \multicolumn{1}{c|}{} & 
        \textit{ZFN} & \textit{ACC} & \textit{AUC} &
        \textit{ZFN} & \textit{ACC} & \textit{AUC} &
        \textit{ZFN} & \textit{ACC} & \textit{AUC} &
        \textit{ZFN} & \textit{ACC} & \textit{AUC} &
        \textit{ZFN} & \textit{ACC} & \textit{AUC} \\

        \hline
        \multirow{3}{*}[1em]
        {\textbf{Hazelnut}} & FID & 98.00 ± 2.14 & 99.14 ± 0.70 & 99.89 ± 0.12 & 74.86 ± 12.89 & 94.57 ± 2.10 & 97.32 ± 0.91 & 51.14 ± 1.67 & 66.57 ± 3.33 & 63.31 ± 4.73 & / & / & / & / & / & / \\ & SSE & 96.29 ± 2.94 & 98.29 ± 1.07 & 99.67 ± 0.25 & 95.43 ± 2.29 & 98.29 ± 0.57 & 99.71 ± 0.14 & 80.29 ± 10.32 & 87.14 ± 7.17 & 92.16 ± 4.90 & 53.43 ± 1.71 & 95.71 ± 0.00 & 92.02 ± 0.29 & 54.29 ± 0.00 & 95.71 ± 0.00 & 92.16 ± 0.00\\ \cline{2-17}
        
        \hline
        \multirow{3}{*}[1em]{\textbf{Screw}} & FID & 52.03 ± 2.18 & 57.63 ± 4.25 & 54.34 ± 8.10 & 50.51 ± 0.68 & 57.46 ± 3.36 & 53.08 ± 5.96 & 51.19 ± 1.15 & 57.63 ± 1.42 & 54.04 ± 3.14 & / & / & / & / & / & / \\
        & SSE & 52.37 ± 3.53 & 57.97 ± 5.46 & 52.81 ± 6.29 & 52.03 ± 3.24 & 57.63 ± 2.89 & 51.31 ± 5.06 & 62.71 ± 10.39 & 77.12 ± 5.22 & 80.86 ± 3.49 & 55.59 ± 4.57 & 66.27 ± 5.42 & 54.02 ± 11.91 & 59.32 ± 0.00 & 90.68 ± 0.00 & 84.83 ± 0.00\\ \cline{2-17}
        
        \hline
        \multirow{3}{*}[1em]{\textbf{Tile}} & FID & 91.43 ± 7.88 & 98.33 ± 1.21 & 99.30 ± 0.73 & 58.81 ± 6.37 & 78.81 ± 1.90 & 83.53 ± 2.61 & 57.62 ± 3.07 & 70.48 ± 2.05 & 72.47 ± 1.56 & / & / & / & / & / & / \\
        & SSE & 78.10 ± 7.85 & 89.76 ± 3.42 & 95.40 ± 2.52 & 52.86 ± 2.21 & 75.24 ± 3.64 & 78.32 ± 2.67 & 50.48 ± 0.58 & 53.57 ± 2.61 & 42.39 ± 5.50 & 61.90 ± 0.00 & 88.10 ± 0.00 & 81.86 ± 0.00 & 61.90 ± 0.00 & 88.10 ± 0.00 & 81.86 ± 0.00 \\ \cline{2-17}
        
        \hline
        \multirow{3}{*}[1em]{\textbf{Wood}} & FID & 91.33 ± 6.78 & 97.00 ± 1.94 & 99.04 ± 0.79 & 71.00 ± 14.85 & 88.67 ± 1.63 & 92.82 ± 2.09 & 53.33 ± 3.80 & 56.00 ± 2.71 & 43.62 ± 3.76 & / & / & / & / & / & / \\
        & SSE & 97.00 ± 3.71 & 97.67 ± 2.91 & 98.89 ± 1.37 & 92.33 ± 6.96 & 96.33 ± 2.87 & 98.48 ± 1.51 & 61.00 ± 4.29 & 71.67 ± 7.67 & 75.09 ± 7.24 & 94.67 ± 10.67 & 95.33 ± 9.33 & 95.02 ± 9.96 & 100.00 ± 0.00 & 100.00 ± 0.00 & 100.00 ± 0.00 \\ \cline{2-17}
        
        \hline
        \hline
        \multirow{3}{*}[1em]{\textbf{Brain MRI}} & FID & 78.57 ± 4.65 & 87.76 ± 3.35 & 93.19 ± 2.56 & 73.27 ± 4.05 & 79.39 ± 2.99 & 80.85 ± 3.92 & 54.90 ± 2.08 & 60.20 ± 3.48 & 58.88 ± 4.83 & / & / & / & / & / & / \\
        & SSE & 84.49 ± 3.95 & 86.94 ± 3.73 & 91.50 ± 2.92 & 84.08 ± 4.81 & 88.37 ± 3.63 & 91.96 ± 3.16 & 61.02 ± 4.63 & 68.37 ± 1.12 & 69.98 ± 2.78 & 50.41 ± 0.50 & 92.45 ± 13.06 & 91.58 ± 12.78 & 50.41 ± 0.50 & 98.98 ± 0.00 & 97.98 ± 0.02\\ \cline{2-17}
        
        \hline
        \multirow{3}{*}[1em]{\textbf{Breast Ultrasound}} & FID & 83.23 ± 6.53 & 91.38 ± 1.23 & 95.45 ± 0.91 & 84.92 ± 3.29 & 85.85 ± 3.46 & 89.62 ± 3.52 & 57.69 ± 2.48 & 70.15 ± 4.83 & 74.41 ± 5.29 & / & / & / & / & / & / \\
        & SSE & 85.23 ± 2.98 & 89.38 ± 3.31 & 92.53 ± 2.41 & 86.46 ± 3.85 & 87.85 ± 3.01 & 91.23 ± 2.31 & 61.23 ± 5.19 & 68.62 ± 2.45 & 71.81 ± 2.29 & 50.62 ± 0.31 & 96.77 ± 4.92 & 97.27 ± 2.42 & 50.62 ± 0.31 & 99.23 ± 0.00 & 98.48 ± 0.01\\ \cline{2-17}
        
        \hline
        \multirow{3}{*}[1em]{\textbf{Retina OCT}} & FID & 50.73 ± 0.59 & 97.23 ± 0.05 & 98.81 ± 0.08 & 50.19 ± 0.08 & 92.10 ± 0.22 & 96.87 ± 0.09 & 50.09 ± 0.07 & 69.92 ± 5.23 & 76.25 ± 6.61 & / & / & / & / & / & / \\
        & SSE & 50.29 ± 0.26 & 96.74 ± 0.07 & 98.49 ± 0.10 & 51.02 ± 0.75 & 96.45 ± 0.04 & 98.33 ± 0.07 & 50.37 ± 0.40 & 79.33 ± 1.34 & 86.86 ± 1.47 & 50.01 ± 0.01 & 93.90 ± 3.03 & 98.24 ± 1.51 & 50.01 ± 0.01 & 99.95 ± 0.05 & 99.97 ± 0.00\\ \cline{2-17}
        
        \hline
        \hline
        \multicolumn{2}{|c|}{\textbf{MEAN}} & \textbf{79.89} & 89.93 & 91.43 & 74.31 & 86.25 & 88.05 & 62.03 & 74.89 & 78.83 & 59.52 & 89.79 & 87.14 & 60.94 & \textbf{96.09} & \textbf{93.61} \\ \cline{2-17}
        
        \hline
    \end{tabular}%
    \end{adjustbox}
    \label{Metric_table}
\end{sidewaystable*}
For CycleGAN-AD-256, CycleGAN-AD-64 and Ganomaly, an anomaly score (SSE or FID) is computed to measure the distance between the original test image and the reconstructed one. As for Padim and PatchCore, only the SSE of the embeddings are considered. Notice that Ganomaly being a GAN method competitor, its anomaly score differs from the one developed by the authors (the difference between the embedding of original and reconstructed images is considered). This is justified by the importance of the reconstructed image quality, which has to bring the pixel-wise difference information needed for the business experts. This is however not possible for the Padim and PatchCore methods that do not generate any reconstructed image. For each methods and datasets, the accuracy (for the FID, when applicable, and SSE anomaly scores) under the zero-false-negative constraint (ZFN columns), and, in a more standard way, without this constraint (ACC columns), as well as the AUCROC (AUC columns) are reported in Table~\ref{Metric_table}. All the metrics are the mean ± the standard deviation (both in percent) over 5 runs for which random train and test splits were generated. Each dataset splits are the same over the methods. The last line gives the mean of each metrics over all the datasets. 

\begin{figure*}
    \centering
    \includegraphics[width=0.8\linewidth]{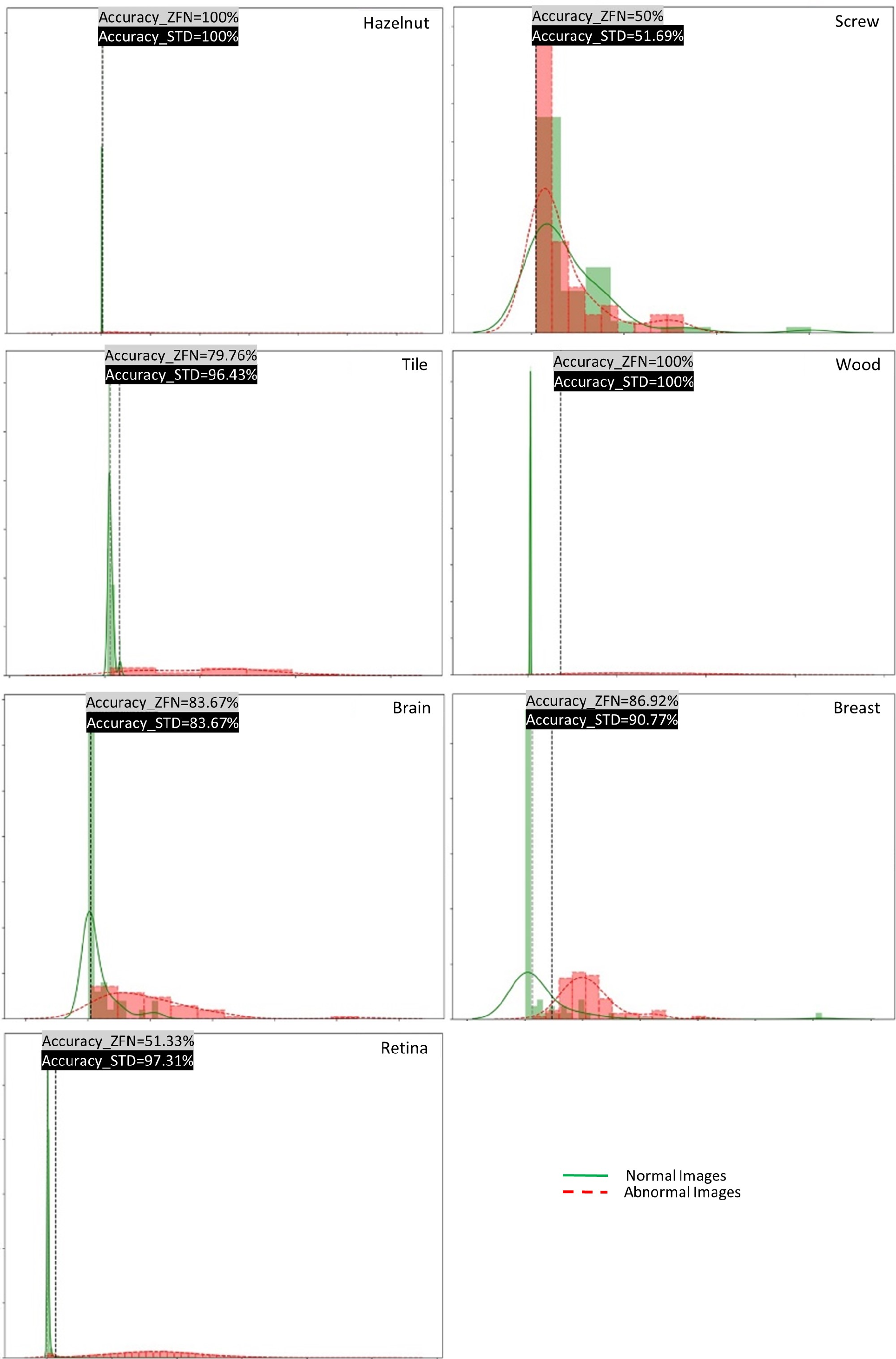}
    
    \caption{(Color online) Anomaly score distributions of normal (solid-green line and bars) and abnormal (dashed-red line and bars) images for the test datasets, with the threshold value in the ZFN setting (vertical dashed line in grey) or in the ACC setting (vertical dashed line in black).}
    \label{fig:Quanti}
\end{figure*}

The anomaly score distributions of the first run for the normal and abnormal test sets of each dataset are shown in Figure~\ref{fig:Quanti}, including the accuracy calculated for the threshold set with the ZFN constraint, or without it.

\section{Discussion}\label{sec:discussion_limitations}

This section discusses the use of Cycle-GAN, for AD with industrial and medical images.

\subsection{Qualitative Discussion}

We observe from the qualitative results presented in Figure~\ref{fig:Quali} that the anomaly reconstruction strongly depends on the nature of the anomaly and the image itself. Indeed, we can see in the red-framed block (abnormal data) that for Hazelnut, Wood and Brain datasets that small cracks, holes or spots are perfectly erased in the reconstructed images, which faithfully highlights the anomalies. These ones present a much higher contrast with the normal image than other datasets like the Screw or Tile ones, where we still observe the anomalies (small scratches or large cracks) after reconstruction. 

Unlike the Tile images where the difference is still well highlighted on the pixelwise difference image, the Screw difference images struggle to highlight the anomaly. For this specific dataset, it can be explained, in addition to the low contrast observed by the anomaly, by the size of the screw in the entire image, which is relatively small compared to the objects or textures considered in the other images. The model correctly reconstructs the background and the shadow, being the majority surface of the image, and the scratches could be interpreted as a reflect instead of an anomaly, in this very small surface. Another explanation could be that the latent space is not small enough to avoid overfitting, preventing to learn efficiently the subtle features in the screw object. It contains too much input information and another architecture could have fixed this issue. On the other hand, the Tile cracks have much less probabilities to be interpreted as an original feature, because it is visually very different that the normal features (colorization, continuity of the texture, etc.).

Good reconstruction is also observed for the Breast Ultrasound or Retina OCT datasets, where the object-shaped anomalies have a large abnormal and well contrasted structure. They are not well erased (specifically for the Breast image) but enough attenuated, which provides sufficient information in the difference image to detect and localize the anomaly. 

We can also observe in the green-framed block (normal data) that the reconstructions (middle images) are more or less identical to the original input (left images), resulting in an almost zero difference image (right images). The model has extracted the features of the normal distributions, and is able to restore normal images without changing the pixels value, thanks to the identity loss. Pixel areas with high discontinuity, as shown in the Screw images, the Brain MRI or the Breast Ultrasound dataset, do not fully follow this observation, resulting in slight differences in the generated image that disturb the anomaly score, and make it difficult to obtain good separable thresholds, in the quantitative step.

From the qualitative results, it is noticeable that the method can reach the anomaly at the pixel level, showing its exact location in the difference images, in spite of the image level labeling the dataset only offers. \\

\subsection{Quantitative Discussion}

For the quantitative assessment, we conclude from Table~\ref{Metric_table} that our method gives an average accuracy of 79.89\% under the zero false negative constraint (ZFN column of CycleGAN-AD-256 method), being at least about 17\% better than other methods in this setup. Without this constraint, it reaches the second position behind the PatchCore method, which is also the case for the AUCROC metric. Notice that even the CycleGAN-AD-64 version is still 12\% better in this setting compared to others. From this statement, we can explain the good results with the ZFN constraint by its ability to reconstruct pretty small or subtle anomalies, still giving a high anomaly score for these challenging images. In other words, unlike other methods, the lowest anomaly score for the abnormal test set is sufficiently distant from the anomaly score population of the normal test set, bringing the threshold so that the false positives are well contained. It means that incorporating abnormal images into the training set, through the cycle consistency framework of Cycle-GAN, helps to better reconstruct the anomalies. This is the case compared to a GAN technique (like Ganomaly) or an embedding technique (like Padim or PatchCore). This observation is particularly interesting for the critical applications in industrial or medical businesses, where this ZFN setup is much more important than the standard accuracy or the AUCROC metric.

In the detail, as expected from the qualitative assessment, the Hazelnut, Tile, Wood, Brain MRI and Breast Ultrasound datasets explain the good average first position of the method, performing better than any other methods. Despite the challenging anomalies that images in Tile or Breast Ultrasound datasets can contain (anomalies not well erased but still sufficiently to be highlighted in the difference images), their abnormal images reach a sufficient anomaly score to be discriminated compared to the normal ones. The Screw and Retina OCT datasets, as for them, reach a quasi 50\% ZFN accuracy, yielding a tool with a quasi zero detection capability, even if the metrics are satisfying for the Retina OCT when we release this constraint (97.23\%). This is typically the sign of the presence of one sample (at least) where the reconstruction struggles, which pushes the ZFN threshold towards a low anomaly score, yielding many false positives. However one can notice that these poor results are also seen in the other methods, where Screw gets its best score of 62.71\% with Ganomaly, and Retina OCT reaches also a near-random discrimination with almost 50\% ZFN accuracy. 

The comparison between our method in a resolution of 256 $\times$ 256 and the same in 64 $\times$ 64 shows the importance of getting higher resolution images, specifically for the industrial datasets (the higher resolution images bring better ZFN accuracy). It seems consistent, when we consider small anomalies consisting of a few pixels or low contrast, that the better the definition, the better the reconstruction and all the processing to find the anomaly score. This statement is not valid for the medical datasets, where the method still performs well at a 64 $\times$ 64 input image resolution. In contrast to the industrial images (Tile is the best example with a drop of about 30\%), the well contrasted anomalies shown in these datasets explain this observation. Another explanation is the presence of larger structures observed in the considered medical images.

For Hazelnut and Tile in the ZFN setup, we can also conclude that the FID anomaly score improves the accuracy by getting rid of the noisy pixel-by-pixel reconstruction described above. For the Wood, the Brain MRI and the Breast Ultrasound datasets, it appears that for the ZFN constraint, the anomaly score based on SSE leads to better results. This could come from the fact that the Inception V3 model is not pre-trained on many medical-like images or highly homogeneous textured images, leading in a poor feature extraction. Depending on the domain, the FID or the SSE anomaly score have to be chosen adequately. However, these scores cannot avoid poor accuracy with datasets like Screw or Retina OCT, where small anomalies are not well captured by the model and the generated images still show the anomalies.

\section{Conclusion}\label{sec:conslusion}
In this work, we propose and characterize for the first time an approach using Cycle-Generative Adversarial Networks (Cycle-GAN) for Anomaly Detection (AD) on industrial and medical images. This method reaches 79.89\% accuracy under a zero false negative constraint, being about 17\% better than other state-of-the-art methods. It exploits the abnormal images at our disposal to refine its representation of normal data, by giving more insights on what is normal or abnormal. Furthermore, thanks to the use of the identity loss, we show that the formalism of Cycle-GAN is naturally well-adapted to perform AD. Particular attention has been given to industrial and medical applications, due to the societal impact it may offer, and motivated by the lack of studies for such kind of work in these areas to date. The proposed method differs from previous work by exploiting both normal and abnormal images to learn mappings that can generate new matched data from one domain to another, under a cycle consistency constraint. The mapping of interest for our AD method is the one that can generate normal images. From this perspective, any differences between the test image and its normal (generated) version can be easily identified. Qualitatively, the pixel squared difference image is used to locate abnormal areas, and then quantitatively, an anomaly score is created to indicate whether the image contains abnormal areas, based on a pre-selected threshold. Ultimately, the method identifies anomalies at the pixel level while the labels are initially at the image level, i.e., without the requirement for tedious annotation at the pixel level.

The achieved results demonstrate that, independent of the application, images presenting anomalies with a sufficient contrast compared to the pixels composing the object or texture considered tend to benefit from higher domain change mapping than those with a low contrast.
For low contrasted anomalies, an exception is observed for images of objects with coarse defects where the localization of anomalies always meet expectations, even with a difficult reconstruction. Issues still remain for object filling a small part of the entire image and with small defects. We argue in this work that when both normal and abnormal data are available for training, the use of Cycle-GAN architectures should be considered as an approach by the community, mainly when the anomalies are known to be of textures or coarse objects.

New applications may also be explored for future work, such as object segmentation or object counting for industrial and medical fields using the same type of cycle-consistent models. This work is a first step and a proof-of-concept for Cycle-GAN in AD for industrial and medical domains.

\section*{Acknowledgment}

V.D. benefits from the support of the Walloon region with a Ph.D. grant from FRIA (F.R.S.-FNRS). M.E. benefits from the support of the Belgian Walloon region for funding SMARTSENS project which is part of Win$^2$WAL program (agreement 2110108). The present research benefited from computational resources made available on Lucia, the Tier-1 supercomputer of the Walloon Region, infrastructure funded by the Walloon Region under the grant agreement n°1910247. The authors thank Charline Dardenne, Jérôme Fink, Géraldin Nanfack and Pierre Poitier for their insightful comments and discussions on this paper.

\bibliographystyle{elsarticle-harv} 
\bibliography{biblio.bib}





\end{document}